\documentclass{article} 
\usepackage{iclr2026_conference,times}

\usepackage{amsmath,amsfonts,bm}

\def\eqref#1{equation~\ref{#1}}

\def\1{\bm{1}}

\DeclareMathAlphabet{\mathsfit}{\encodingdefault}{\sfdefault}{m}{sl}
\SetMathAlphabet{\mathsfit}{bold}{\encodingdefault}{\sfdefault}{bx}{n}

\usepackage{hyperref}
\usepackage{url}
\usepackage{booktabs}
\usepackage{graphicx}
\usepackage{hyperref}

\title{Sat-JEPA-Diff: Bridging Self-Supervised Learning and Generative Diffusion for Remote Sensing}

\author{Kürşat Kömürcü, Linas Petkevicius \thanks{This project was funded by the European Union (project No S-MIP-23-45) under the agreement with the Research Council of Lithuania (LMTLT).} \\
Vilnius University\\
Faculty of Mathematics and Informatics\\
Institute of Computer Science\\ 
Artificial Intelligence Methods Lab\\
Vilnius, LT-03225, Lithuania \\
\texttt{\{kursat.komurcu,linas.petkevicius\}@mif.vu.lt} \\
}

\iclrfinalcopy 
\begin{document}

\maketitle

\begin{abstract}
Predicting satellite imagery requires a balance between structural accuracy and textural detail. Standard deterministic methods like PredRNN or SimVP minimize pixel-based errors but suffer from the "regression to the mean" problem, producing blurry outputs that obscure subtle geographic-spatial features. Generative models provide realistic textures but often misleadingly reveal structural anomalies. To bridge this gap, we introduce Sat-JEPA-Diff, which combines Self-Supervised Learning (SSL) with Hidden Diffusion Models (LDM). An IJEPA module predicts stable semantic representations, which then route a frozen Stable Diffusion backbone via a lightweight cross-attention adapter. This ensures that the synthesized high-accuracy textures are based on absolutely accurate structural predictions. Evaluated on a global Sentinel-2 dataset, Sat-JEPA-Diff excels at resolving sharp boundaries. It achieves leading perceptual scores (GSSIM: 0.8984, FID: 0.1475) and significantly outperforms deterministic baselines, despite standard autoregressive stability limits. The code and dataset are publicly available on \href{https://github.com/VU-AIML/SAT-JEPA-DIFF}{GitHub}.
\end{abstract}

\section{Introduction}
Continuous Earth observation is crucial for environmental monitoring but hindered by frequent cloud cover. Therefore, spatiotemporal forecasting ($t \to t+1$) becomes an indispensable "virtual sensor" that helps bridge the gap. Current forecasting models face a fundamental trade-off. Deterministic methods, such as PredRNN \cite{wang2022predrnn} and SimVP \cite{Gao2022}, focus on optimizing pixel-wise error (MSE). This, however, results in "regression toward the mean" with blurry images that lack spectral detail. On the contrary, generative forecasting approaches, like Denoising Diffusion Probabilistic Models (DDPMs) \cite{Rombach2022}, excel in reproducing plausible textures. Nevertheless, these are often "hallucinations" that generate plausible but false structures, especially without adequate semantic guidance.

In this paper, we propose Sat-JEPA-Diff, a novel spatiotemporal forecasting model that integrates the benefits of self-supervised learning with generative capabilities. We adapt IJEPA \cite{Assran2023} to forecast pre-computed Alpha Earth Foundation Model \cite{brown2025alphaearth} embeddings, providing robust SOTA semantic guidance to a frozen Latent Diffusion Model via a custom cross-attention adapter. Details on the dataset are provided in Appendix~\ref{app:dataset} and Figure~\ref{fig:study_map}.

\section{Related Work}

\textbf{Spatiotemporal Forecasting.} Early studies used RNNs like ConvLSTM \cite{shi2015convolutional} and PredRNN \cite{wang2022predrnn} for temporal modeling, but these are computationally intensive. CNN-based models like SimVP \cite{Gao2022} improved efficiency via spatial convolutions. However, these deterministic models optimize pixel-wise L1/L2 losses, biasing predictions towards the mean and producing blurry outputs lacking high-frequency detail \cite{Mathieu2015}.

\textbf{Generative Models in Remote Sensing.} While GANs \cite{Goodfellow2014} mitigate blurring, they suffer from mode collapse. Denoising Diffusion Probabilistic Models (DDPMs) and Latent Diffusion Models (LDMs) have emerged as promising alternatives to GANs for texture synthesis tasks. However, using LDMs for satellite time-series images is not an easy task, as these models are prone to "hallucinating" features such as road topologies or buildings, etc., without any semantic constraints \cite{Saharia2022}.

\textbf{Self-Supervised Representation Learning.} Effective structure-texture bridging requires semantic guidance. While MAE \cite{He2022} and SatMAE \cite{Cong2022} learn representations by reconstructing pixels, they overfit on noise. IJEPA \cite{Assran2023} avoids this by learning in abstract space. Concurrently, geo foundation models such as Alpha Earth \cite{brown2025alphaearth}, Panopticon \cite{waldmann2025panopticon}, and TerraMind \cite{jakubik2025terramind} provide powerful pre-trained EO embeddings. Sat-JEPA-Diff leverages IJEPA with such embeddings for diffusion conditioning, avoiding spatial collapse of prior hybrid approaches.

\section{Methodology}

\begin{figure}[t]
\centering
\includegraphics[width=\linewidth]{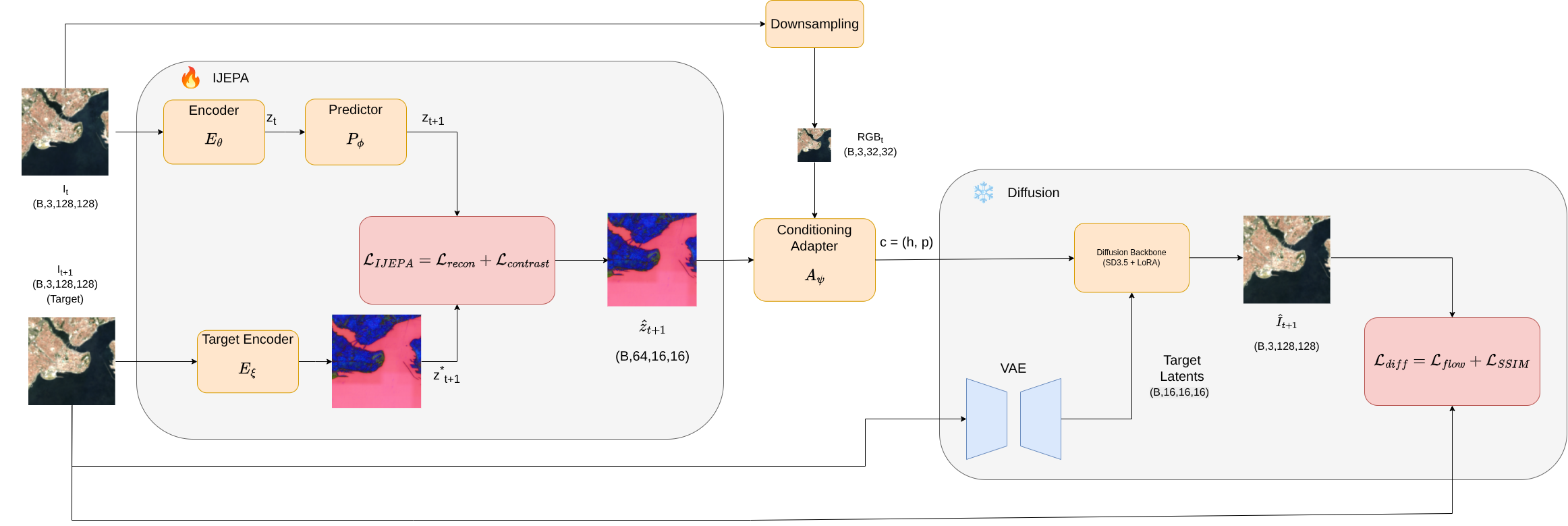}
\caption{Overview of Sat-JEPA-Diff. The IJEPA module (left) predicts future semantic embeddings $\hat{z}_{t+1}$ from input $I_t$. These embeddings, combined with coarse spatial structure, condition a frozen SD3.5 backbone via a learned adapter to generate $\hat{I}_{t+1}$.}
\label{fig:architecture}
\end{figure}

Our framework, illustrated in Figure~\ref{fig:architecture}, consists of two synergistic modules: (1) an IJEPA-based temporal predictor that forecasts future semantic embeddings, and (2) a conditioned diffusion generator that synthesizes high-fidelity RGB imagery guided by these predictions. Full architectural specifications and hyperparameters are provided in Appendix~\ref{app:implementation}.

\textbf{Problem Formulation.} Given a satellite image $I_t \in \mathbb{R}^{3 \times H \times W}$ at time $t$, our goal is to predict the corresponding image $I_{t+1}$ at time $t+1$. We decompose this into two stages: (i) predicting the semantic representation $\hat{z}_{t+1}$ of the future frame, and (ii) generating the RGB output conditioned on this prediction.

\textbf{IJEPA Temporal Prediction.} We adapt the Joint-Embedding Predictive Architecture (IJEPA)~\cite{Assran2023} for temporal forecasting. Unlike masked autoencoders that reconstruct pixels, IJEPA operates entirely in latent space, learning representations robust to sensor noise and atmospheric variations.

\textbf{Encoder.} A Vision Transformer $E_\theta$ processes the input image $I_t$ into patch embeddings:
\begin{equation*}
z_t = E_\theta(I_t) \in \mathbb{R}^{N \times D}
\end{equation*}
where $N = (H/p)^2$ is the number of patches with patch size $p$, and $D$ is the embedding dimension.

\textbf{Predictor.} A transformer-based predictor $P_\phi$ forecasts future embeddings from the encoded representation:
\begin{equation*}
\hat{z}_{t+1} = P_\phi(z_t) \in \mathbb{R}^{N \times D}
\end{equation*}

\textbf{Target Encoder.} Following IJEPA, we maintain an exponential moving average (EMA) copy $E_\xi$ of the encoder to produce stable target embeddings $z^*_{t+1} = E_\xi(I_{t+1})$.

\textbf{IJEPA Loss.} We employ a hybrid loss combining reconstruction and contrastive objectives (see Appendix~\ref{app:loss_justification} for component justification):
\begin{equation*}
\mathcal{L}_{\text{IJEPA}} = \lambda_1 \|\hat{z}_{t+1} - z^*_{t+1}\|_1 + \lambda_2 (1 - \cos(\hat{z}_{t+1}, z^*_{t+1})) + \lambda_3 \mathcal{L}_{\text{spatial}} + \lambda_4 \mathcal{L}_{\text{contrast}}
\end{equation*}
where $\mathcal{L}_{\text{spatial}}$ penalizes variance mismatch between predicted and target embeddings to prevent spatial collapse, and $\mathcal{L}_{\text{contrast}}$ is an InfoNCE loss ensuring global discriminability.

\textbf{Conditioned Diffusion Generation.} We leverage Stable Diffusion 3.5~\cite{esser2024scaling} as our generative backbone, keeping the core transformer frozen and training only a lightweight conditioning adapter with LoRA~\cite{hu2022lora}.

\textbf{Conditioning Adapter.} The adapter $A_\psi$ transforms IJEPA embeddings into cross-attention conditioning signals:
\begin{equation*}
c = (h, p) = A_\psi(\hat{z}_{t+1}, I_t^c)
\end{equation*}
where $h \in \mathbb{R}^{M \times 4096}$ provides token-level conditioning via cross-attention, $p \in \mathbb{R}^{2048}$ provides global conditioning, and $I_t^c$ is a coarse $32 \times 32$ downsampled version of $I_t$ that preserves low-frequency spatial structure.

The adapter employs a learned fusion gate $\alpha$ to balance semantic (IJEPA) and structural (coarse RGB) signals:
\begin{equation*}
h = \alpha \cdot h_{\text{semantic}} + (1 - \alpha) \cdot h_{\text{coarse}}
\end{equation*}

\textbf{Flow Matching Objective.} Following rectified flow formulation~\cite{liu2022flow}, we train with velocity prediction. Given target latents $x_0$ from VAE-encoded $I_{t+1}$ and noise $\epsilon \sim \mathcal{N}(0, I)$, we construct:
\begin{equation*}
x_\sigma = (1 - \sigma) x_0 + \sigma \epsilon, \quad v^* = \epsilon - x_0
\end{equation*}
The diffusion backbone learns to predict this velocity:
\begin{equation*}
\mathcal{L}_{\text{diff}} = \|v_\theta(x_\sigma, \sigma, c) - v^*\|^2 + \lambda_{\text{ssim}} \mathcal{L}_{\text{SSIM}}
\end{equation*}

\textbf{Total Objective.} The complete training loss combines both modules:
\begin{equation*}
\mathcal{L} = \mathcal{L}_{\text{IJEPA}} + \lambda \mathcal{L}_{\text{diff}}
\end{equation*}

\section{Results}

\begin{table}[h]
\centering
\caption{Quantitative comparison on the test set. Arrows indicate whether lower ($\downarrow$) or higher ($\uparrow$) values are better. Best results are highlighted in \textbf{bold}.}
\label{tab:results}
\small
\setlength{\tabcolsep}{3pt} 
\resizebox{\linewidth}{!}{%
\begin{tabular}{lccccccc}
\toprule
\textbf{Model} & \textbf{L1} $\downarrow$ & \textbf{MSE} $\downarrow$ & \textbf{PSNR} $\uparrow$ & \textbf{SSIM} $\uparrow$ & \textbf{GSSIM} $\uparrow$ & \textbf{LPIPS} $\downarrow$ & \textbf{FID} $\downarrow$ \\
\midrule
\multicolumn{8}{l}{\textit{Deterministic Baselines}} \\
Default & 0.0131 & 0.0008 & 37.52 & 0.9361 & 0.7858 & \textbf{0.0708} & 0.6959 \\
PredRNN \cite{wang2022predrnn} & \textbf{0.0117} & 0.0005 & \textbf{38.38} & \textbf{0.9476} & 0.7836 & 0.0726 & 9.9720 \\
SimVP v2 \cite{Gao2022} & 0.0131 & 0.0006 & 37.63 & 0.9391 & 0.7719 & 0.0928 & 18.7208 \\
\midrule
\multicolumn{8}{l}{\textit{Generative Models}} \\
SD 3.5 \cite{esser2024scaling} & 0.0175 & 0.0005 & 32.98 & 0.8398 & 0.8711 & 0.4528 & 0.1533 \\
MCVD \cite{voleti2022mcvd} & 0.0314 & 0.0031 & 31.28 & 0.8637 & 0.7665 & 0.1890 & 0.1956 \\
\textbf{Ours (Panopticon)} \cite{waldmann2025panopticon} & 0.0179 & 0.0005 & 32.89 & 0.8398 & 0.8750 & 0.4475 & 0.1475 \\
\textbf{Ours} & 0.0158 & \textbf{0.0004} & 33.81 & 0.8672 & \textbf{0.8984} & 0.4449 & \textbf{0.1475} \\
\bottomrule
\end{tabular}%
}
\end{table}

As presented in Table \ref{tab:results}, deterministic baselines (PredRNN, SimVP) achieve high PSNR and SSIM scores. However, these metrics are pixel-wise and encourage blurred predictions that "regress towards the mean." In contrast, Sat-JEPA-Diff achieves a significantly higher GSSIM score of 0.8984, which is an increase of over 11\% over the best baseline. GSSIM evaluates the preservation of edges and structural gradients. Our results indicate that semantic guidance successfully maintains sharp geospatial features such as roads and urban areas avoiding the blurring typical of deterministic models. Furthermore, swapping the ViT encoder for Panopticon \cite{waldmann2025panopticon} produces similar perceptual results, demonstrating that our approach is not dependent on a single encoder type. Although SSIM drops slightly, this decrease reflects the well-known perception-distortion trade-off \cite{blau2018perception}, as the model prioritizes realistic texture synthesis over exact pixel averaging.

\begin{figure}[t]
\centering
\includegraphics[width=0.8\linewidth]{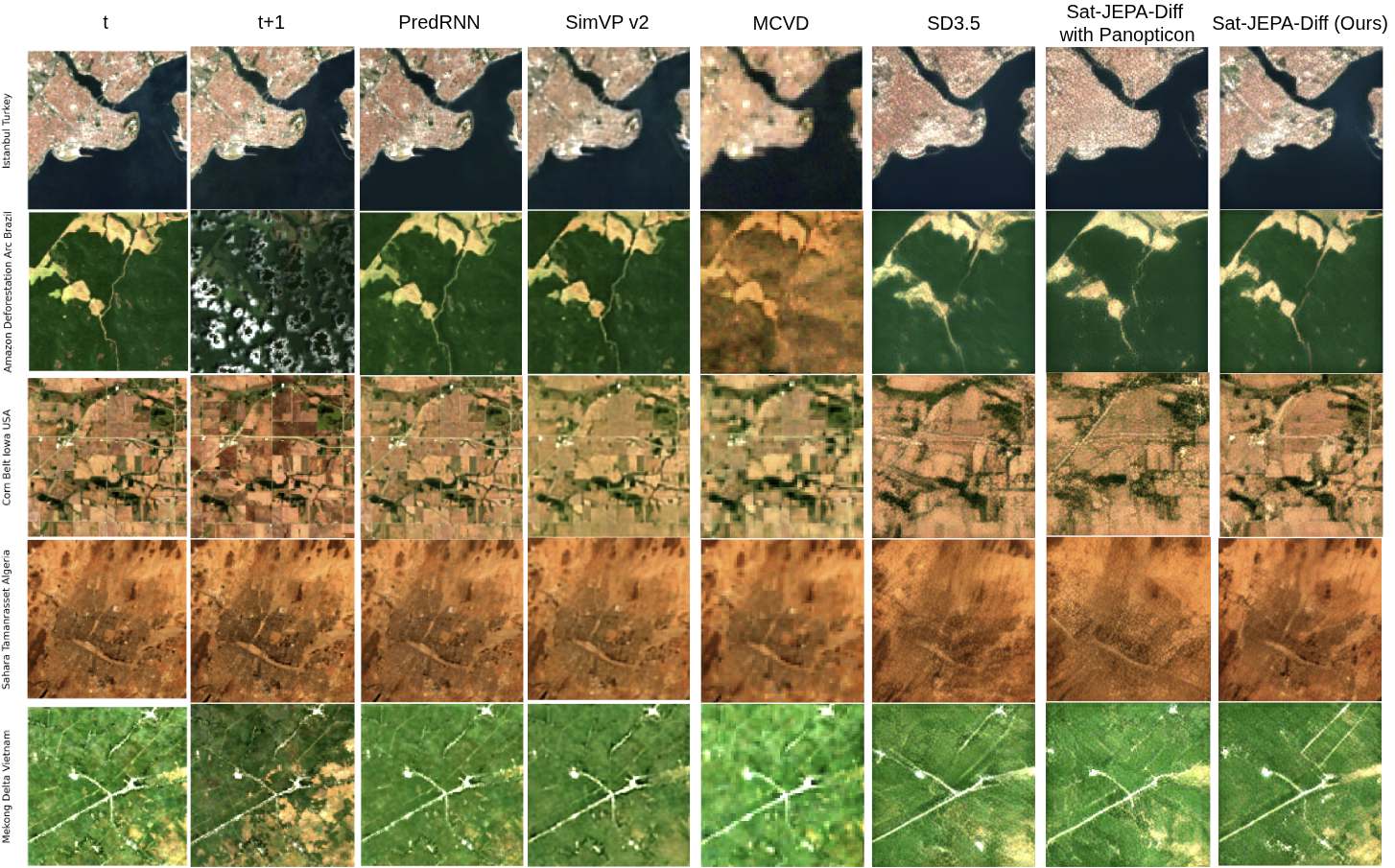}
\caption{Qualitative comparison of next-frame predictions ($t \to t+1$). While deterministic baselines (PredRNN, SimVP) suffer from spectral blurring, Sat-JEPA-Diff preserves high-frequency details and geospatial boundaries.}
\label{fig:example_images}
\end{figure}

Qualitative Analysis. Figure \ref{fig:example_images} presents a visual comparison of next-frame predictions ($t \to t+1$). Deterministic baselines like PredRNN and SimVP suffer from spectral blurring, masking Istanbul's urban density and smearing Amazonian agricultural borders. Sat-JEPA-Diff overcomes this by generating crisp, realistic textures. Our model preserves intricate street networks and distinct forest edges, validated by our superior GSSIM scores. While diffusion-based generation introduces minor stochastic variations, the semantic grounding provided by the IJEPA module ensures these variations remain geosemantically plausible. We further validate long-horizon temporal consistency through autoregressive rollout experiments in Appendix~\ref{app:rollout}.

\section{Conclusion}
We presented Sat-JEPA-Diff, a novel framework integrating the semantic reasoning of IJEPA with the generative power of Latent Diffusion Models. By shifting forecasting from pixel space to a semantic latent space, we successfully mitigate the blurring artifacts of deterministic baselines. Our results demonstrate superior performance in structural integrity (GSSIM) and perceptual quality (FID). Future work will explore replacing learned embeddings with vision-language scene descriptions as temporally forecastable conditioning signals for generation.

\bibliography{iclr2026_conference}

@article{Assran2023,
  title={Self-supervised learning from images with a joint-embedding predictive architecture},
  author={Assran, Mahmoud and Duval, Quentin and Misra, Ishan and Bojanowski, Piotr and Vincent, Pascal and Rabbat, Michael and LeCun, Yann and Ballas, Nicolas},
  journal={Proceedings of the IEEE/CVF Conference on Computer Vision and Pattern Recognition},
  pages={15619--15629},
  year={2023}
}

@inproceedings{Rombach2022,
  title={High-resolution image synthesis with latent diffusion models},
  author={Rombach, Robin and Blattmann, Andreas and Lorenz, Dominik and Esser, Patrick and Ommer, Bj{\"o}rn},
  booktitle={Proceedings of the IEEE/CVF conference on computer vision and pattern recognition},
  pages={10684--10695},
  year={2022}
}

@inproceedings{Gao2022,
  title={Simvp: Simpler is better for video prediction},
  author={Gao, Zhangyang and Tan, Cheng and Wu, Lirong and Li, Stan Z},
  booktitle={Proceedings of the IEEE/CVF Conference on Computer Vision and Pattern Recognition},
  pages={3950--3959},
  year={2022}
}

@article{He2022,
  title={Masked autoencoders are scalable vision learners},
  author={He, Kaiming and Chen, Xinlei and Xie, Saining and Li, Yanghao and Doll{\'a}r, Piotr and Girshick, Ross},
  journal={Proceedings of the IEEE/CVF conference on computer vision and pattern recognition},
  pages={16000--16009},
  year={2022}
}

@article{wang2022predrnn,
  title={Predrnn: A recurrent neural network for spatiotemporal predictive learning},
  author={Wang, Yunbo and Wu, Haixu and Zhang, Jianjin and Gao, Zhifeng and Wang, Jianmin and Yu, Philip S and Long, Mingsheng},
  journal={IEEE Transactions on Pattern Analysis and Machine Intelligence},
  volume={45},
  number={2},
  pages={2208--2225},
  year={2022},
  publisher={IEEE}
}

@article{Mathieu2015,
  title={Deep multi-scale video prediction beyond mean square error},
  author={Mathieu, Michael and Couprie, Camille and LeCun, Yann},
  journal={arXiv preprint arXiv:1511.05440},
  year={2015}
}

@article{Saharia2022,
  title={Image super-resolution via iterative refinement},
  author={Saharia, Chitwan and Ho, Jonathan and Chan, William and Salimans, Tim and Fleet, David J and Norouzi, Mohammad},
  journal={IEEE Transactions on Pattern Analysis and Machine Intelligence},
  volume={45},
  number={4},
  pages={4713--4726},
  year={2022}
}

@article{Goodfellow2014,
  title={Generative adversarial nets},
  author={Goodfellow, Ian J and Pouget-Abadie, Jean and Mirza, Mehdi and Xu, Bing and Warde-Farley, David and Ozair, Sherjil and Courville, Aaron and Bengio, Yoshua},
  journal={Advances in neural information processing systems},
  volume={27},
  year={2014}
}

@inproceedings{Cong2022,
 author = {Cong, Yezhen and Khanna, Samar and Meng, Chenlin and Liu, Patrick and Rozi, Erik and He, Yutong and Burke, Marshall and Lobell, David and Ermon, Stefano},
 booktitle = {Advances in Neural Information Processing Systems},
 pages = {197--211},
 publisher = {Curran Associates, Inc.},
 title = {SatMAE: Pre-training Transformers for Temporal and Multi-Spectral Satellite Imagery},
 volume = {35},
 year = {2022}
}

@article{voleti2022mcvd,
  title={Mcvd-masked conditional video diffusion for prediction, generation, and interpolation},
  author={Voleti, Vikram and Jolicoeur-Martineau, Alexia and Pal, Chris},
  journal={Advances in neural information processing systems},
  volume={35},
  pages={23371--23385},
  year={2022}
}

@inproceedings{esser2024scaling,
  title={Scaling rectified flow transformers for high-resolution image synthesis},
  author={Esser, Patrick and Kulal, Sumith and Blattmann, Andreas and Entezari, Rahim and M{\"u}ller, Jonas and Saini, Harry and Levi, Yam and Lorenz, Dominik and Sauer, Axel and Boesel, Frederic and others},
  booktitle={Forty-first international conference on machine learning},
  year={2024}
}

@article{liu2022flow,
  title={Flow straight and fast: Learning to generate and transfer data with rectified flow},
  author={Liu, Xingchao and Gong, Chengyue and Liu, Qiang},
  journal={arXiv preprint arXiv:2209.03003},
  year={2022}
}

@article{shi2015convolutional,
  title={Convolutional LSTM network: A machine learning approach for precipitation nowcasting},
  author={Shi, Xingjian and Chen, Zhourong and Wang, Hao and Yeung, Dit-Yan and Wong, Wai-Kin and Woo, Wang-chun},
  journal={Advances in neural information processing systems},
  volume={28},
  year={2015}
}

@inproceedings{blau2018perception,
  title={The perception-distortion tradeoff},
  author={Blau, Yochai and Michaeli, Tomer},
  booktitle={Proceedings of the IEEE conference on computer vision and pattern recognition},
  pages={6228--6237},
  year={2018}
}

@article{hu2022lora,
  title={Lora: Low-rank adaptation of large language models.},
  author={Hu, Edward J and Shen, Yelong and Wallis, Phillip and Allen-Zhu, Zeyuan and Li, Yuanzhi and Wang, Shean and Wang, Lu and Chen, Weizhu and others},
  journal={ICLR},
  volume={1},
  number={2},
  pages={3},
  year={2022}
}

@article{brown2025alphaearth,
  title={Alphaearth foundations: An embedding field model for accurate and efficient global mapping from sparse label data},
  author={Brown, Christopher F and Kazmierski, Michal R and Pasquarella, Valerie J and Rucklidge, William J and Samsikova, Masha and Zhang, Chenhui and Shelhamer, Evan and Lahera, Estefania and Wiles, Olivia and Ilyushchenko, Simon and others},
  journal={arXiv preprint arXiv:2507.22291},
  year={2025}
}

@inproceedings{waldmann2025panopticon,
  title={Panopticon: Advancing Any-Sensor Foundation Models for Earth Observation},
  author={Waldmann, Leonard and Shah, Ando and Wang, Yi and Lehmann, Nils and Stewart, Adam J. and Xiong, Zhitong and Zhu, Xiao Xiang and Bauer, Stefan and Chuang, John},
  booktitle={Proceedings of the IEEE/CVF Conference on Computer Vision and Pattern Recognition Workshops (CVPRW)},
  year={2025}
}

@article{jakubik2025terramind,
  title={TerraMind: Large-Scale Generative Multimodality for Earth Observation},
  author={Jakubik, Johannes and Yang, Felix and Blumenstiel, Benedikt and Scheurer, Erik and Sedona, Rocco and Maurogiovanni, Stefano and Bosmans, Jente and Dionelis, Nikolaos and Marsocci, Valerio and Kopp, Niklas and Ramachandran, Rahul and Fraccaro, Paolo and Brunschwiler, Thomas and Cavallaro, Gabriele and Bernab{\'e}-Moreno, Juan and Long{\'e}p{\'e}, Nicolas},
  journal={IEEE/CVF International Conference on Computer Vision (ICCV)},
  year={2025}
}
\bibliographystyle{iclr2026_conference}


\appendix
\section{Dataset Details}\label{app:dataset}

To train and evaluate the Sat-JEPA-Diff architecture, we curated a large-scale, multi-modal dataset spanning diverse geographical landscapes.

\subsection{Data Sources}
\begin{itemize}
    \item \textbf{Embeddings:} Alpha Earth Foundation Model Embeddings. We use pre-computed 64-dimensional feature vectors per pixel, which contain semantic information resistant to noise from the atmosphere.
\item \textbf{Optical Imagery:} Sentinel-2 Surface Reflectance (RGB bands), harmonized to 10m GSD. Images are normalized to the range [0,1].
\end{itemize}

\subsection{Spatiotemporal Distribution}
The data set covers a period from 2017 to 2024, and there are 100 unique Regions of Interest (RoIs).

\begin{figure}[h]
\centering
\includegraphics[width=0.8\linewidth]{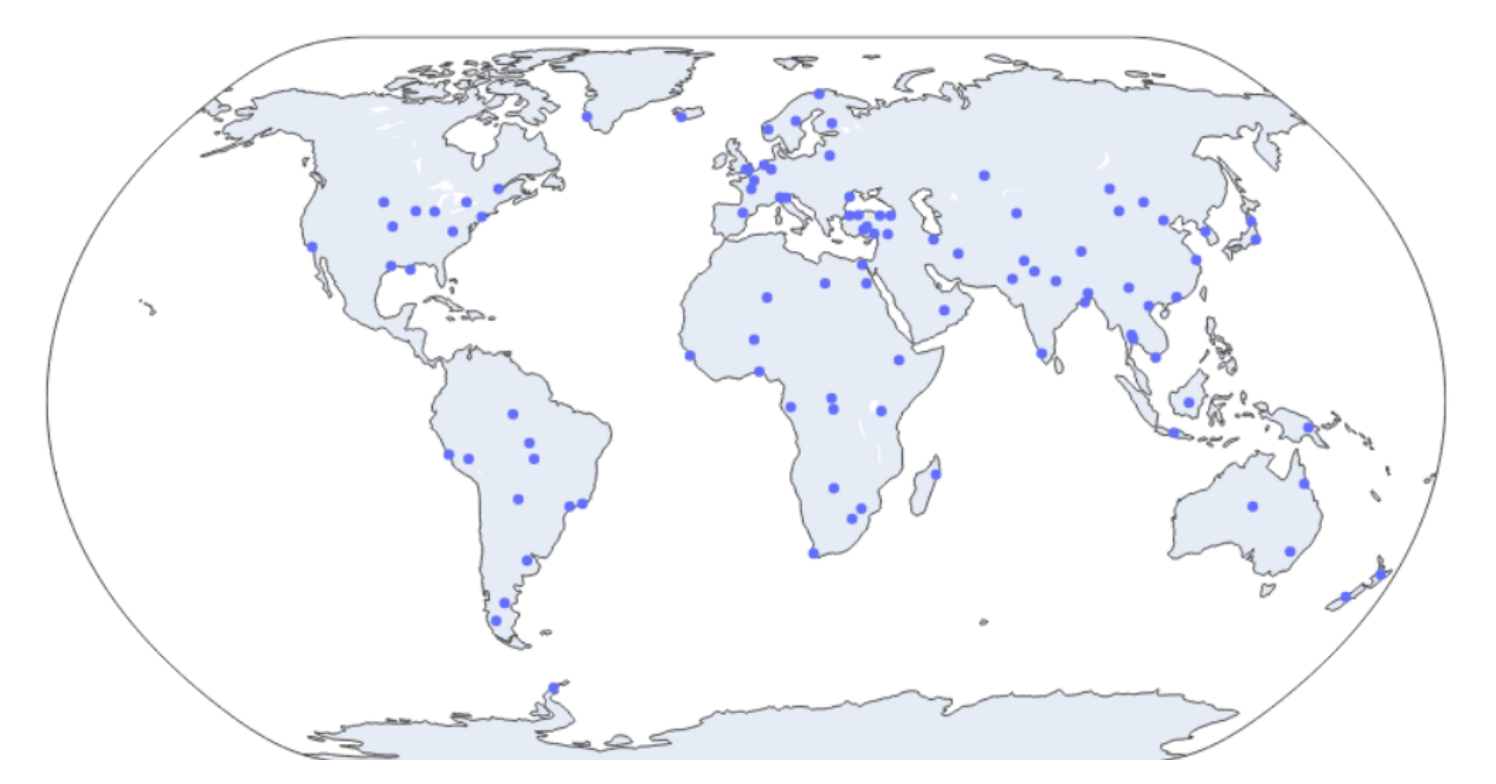} 
\caption{Geographical distribution of the 100 selected Regions of Interest (RoIs).}
\label{fig:study_map}
\end{figure}

\section{Implementation Details}\label{app:implementation}

Our architectural configurations, hyperparameters, and training infrastructure are summarized below.

\subsection{Network Architecture}

\textbf{IJEPA Module.} We use a Vision Transformer (ViT) backbone as the encoder $E_\theta$. The encoder processes $128 \times 128$ RGB input images with patch size $8$, producing $N = 256$ patch tokens. The embedding dimension is $D = 768$ for the base configuration. The predictor $P_\phi$ is a lightweight transformer with depth configurable via hyperparameter (default: 6 layers). The target encoder $E_\xi$ is an exponential moving average (EMA) copy of $E_\theta$, updated with momentum $\tau \in [0.999, 1.0]$ following a cosine schedule.

\textbf{Projection Head.} A linear projection layer maps the predictor's raw output from $768$ dimensions to the target embedding dimension of $64$, matching the Alpha Earth Foundation Model embedding space used as supervision signal.

\textbf{Conditioning Adapter.} The adapter $A_\psi$ transforms IJEPA embeddings and coarse RGB signals into SD3.5-compatible conditioning. It consists of:
\begin{itemize}
    \item \textbf{IJEPA Token Projection:} A three-layer MLP ($768 \to 1024 \to 1024 \to 4096$) with LayerNorm and GELU activations, projecting semantic tokens to the cross-attention dimension.
    \item \textbf{Coarse RGB Projection:} The $32 \times 32$ reference image is patchified into $8 \times 8 = 64$ tokens (patch size $4 \times 4$), each with dimension $48$ (flattened $4 \times 4 \times 3$ RGB values). A similar MLP ($48 \to 1024 \to 1024 \to 4096$) projects these to cross-attention space.
    \item \textbf{Pooled Projection:} Mean-pooled IJEPA tokens are projected via a two-layer MLP ($768 \to 1024 \to 2048$) for global conditioning.
    \item \textbf{Fusion Gate:} A learned sigmoid gate balances IJEPA semantic signals and coarse RGB structural signals: $h = \alpha \cdot h_{\text{semantic}} + (1 - \alpha) \cdot h_{\text{coarse}}$.
\end{itemize}
Learnable positional embeddings (up to 1024 tokens for IJEPA, 64 tokens for coarse RGB) are added before projection. Total adapter parameters: $\sim$25M.

\textbf{Stable Diffusion Backbone.} We use Stable Diffusion 3.5 Medium~\cite{esser2024scaling} as the generative backbone. The core transformer is kept frozen, with only Low-Rank Adaptation (LoRA)~\cite{hu2022lora} modules trained on the attention projections. LoRA configuration: rank $r = 8$, alpha $\alpha = 16$. The VAE encoder/decoder operates at $8\times$ spatial compression.

\subsection{Training Configuration}

\textbf{Optimization.} We use AdamW optimizer with the following schedule:
\begin{itemize}
    \item Base learning rate: $1 \times 10^{-4}$
    \item Start learning rate: $2 \times 10^{-5}$ (linear warmup)
    \item Final learning rate: $1 \times 10^{-6}$ (cosine decay)
    \item Warmup epochs: 1
    \item Weight decay: $0.04 \to 0.4$ (cosine schedule)
    \item Total epochs: 100
\end{itemize}

\textbf{Mixed Precision.} Training uses \texttt{bfloat16} automatic mixed precision for memory efficiency. The VAE is kept in \texttt{float32} for numerical stability during encoding/decoding operations.

\textbf{Batch Size.} We trained the model with a batch size of 8 per GPU. For multi-GPU training, we use PyTorch DistributedDataParallel with synchronized batch normalization.

\textbf{EMA Schedule.} Target encoder momentum follows: $\tau_t = \tau_{\text{base}} + t \cdot (\tau_{\text{final}} - \tau_{\text{base}}) / T$, where $\tau_{\text{base}} = 0.999$, $\tau_{\text{final}} = 1.0$, and $T$ is the total training iterations.

\subsection{Loss Function Weights}

\textbf{IJEPA Hybrid Loss.} We employ a multi-component loss to prevent spatial collapse:
\begin{itemize}
    \item L1 reconstruction weight ($\lambda_1$): $20.0$
    \item Cosine similarity weight ($\lambda_2$): $2.0$
    \item Spatial variance weight ($\lambda_3$): $2.0$
    \item Contrastive (InfoNCE) weight ($\lambda_4$): $0.5$
    \item Feature regression weight: $5.0$
    \item Contrastive temperature: $\tau = 0.1$
\end{itemize}

\textbf{Diffusion Loss.} Flow matching MSE loss with SSIM regularization (weight $0.1$). Total loss combines IJEPA and diffusion objectives with SD loss weight $\lambda = 1.0$.

\textbf{Reference Dropout.} During training, the coarse RGB reference is dropped with probability $p = 0.15$ to encourage the model to rely on IJEPA semantic predictions alone, improving generalization.

\subsection{Masking Strategy}

Following IJEPA~\cite{Assran2023}, we use multi-block masking:
\begin{itemize}
    \item Encoder mask scale: $[0.7, 1.0]$ (proportion of patches visible)
    \item Predictor mask scale: $[0.2, 0.5]$ (proportion of patches to predict)
    \item Aspect ratio range: $[0.75, 1.5]$
    \item Number of encoder masks: 1
    \item Number of predictor masks: 1
    \item Minimum patches to keep: 6
    \item Overlap allowed: True
\end{itemize}

\subsection{Inference Configuration}

\textbf{Diffusion Sampling.} At inference, we use the Flow Matching Euler Discrete scheduler with:
\begin{itemize}
    \item Default sampling steps: 20
    \item Noise strength for latent initialization: $\sigma = 0.35$
    \item Single-step speed estimation to increase efficiency.
\end{itemize}

\textbf{Conditioning Fusion.} The $32 \times 32$ RGB image $I_t$, with a reduced sampling rate, captures large spatial patterns, while the IJEPA embedded vectors $z_{t+1}$ provide detailed semantic cues. During training, the merging gate stabilizes around $\alpha \approx 0.5$, indicating that the network treats both inputs with equal weight.

\subsection{Computational Resources}

\textbf{Hardware.} All experiments were performed on a single NVIDIA RTX 5090 24GB GPU.

\subsection{Text Conditioning}

We use a fixed prompt for all Sentinel-2 imagery:
\begin{quote}
\textit{``High-resolution Sentinel-2 satellite image, multispectral earth observation, natural colors RGB composite, 10m ground resolution, clear atmospheric conditions, detailed land surface features''}
\end{quote}

\subsection{Ablation on IJEPA Loss Components (Training Dynamics)}\label{app:loss_justification}

\begin{figure}[t]
\centering
\includegraphics[width=0.75\linewidth]{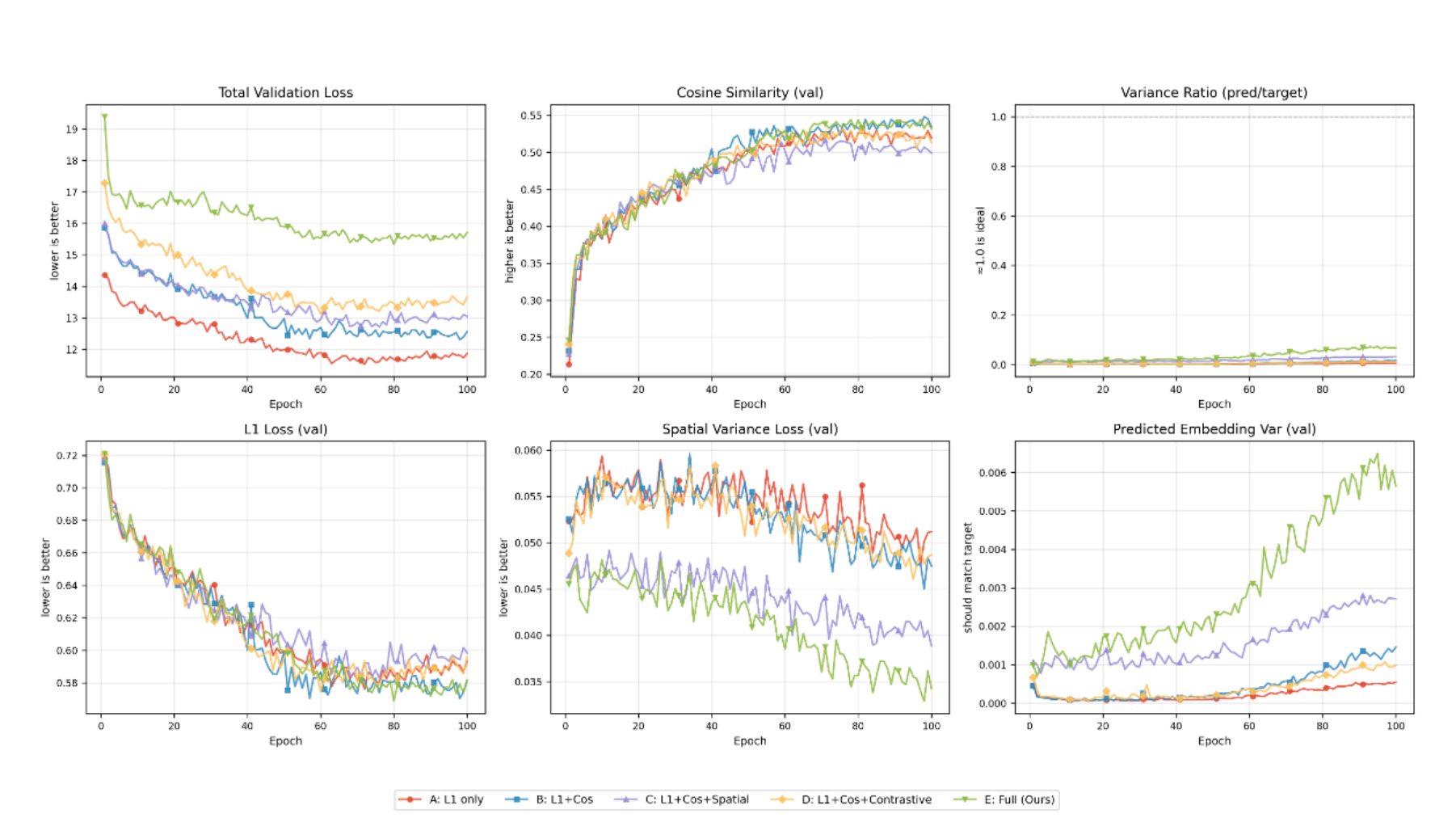} 
\caption{Systematic IJEPA Loss Ablation. Validation metrics over 100 epochs demonstrate that our full objective function (E curve) uniquely avoids representation collapse and maintains high embedding variance compared to near-zero variance in the reduced underlying models (A-D curves). Despite higher total loss, the full model maintains high cosine similarity and achieves superior spatial variance.}
\label{fig:loss_ablation}
\end{figure}

\begin{figure}[h]
\centering
\includegraphics[width=0.8\linewidth]{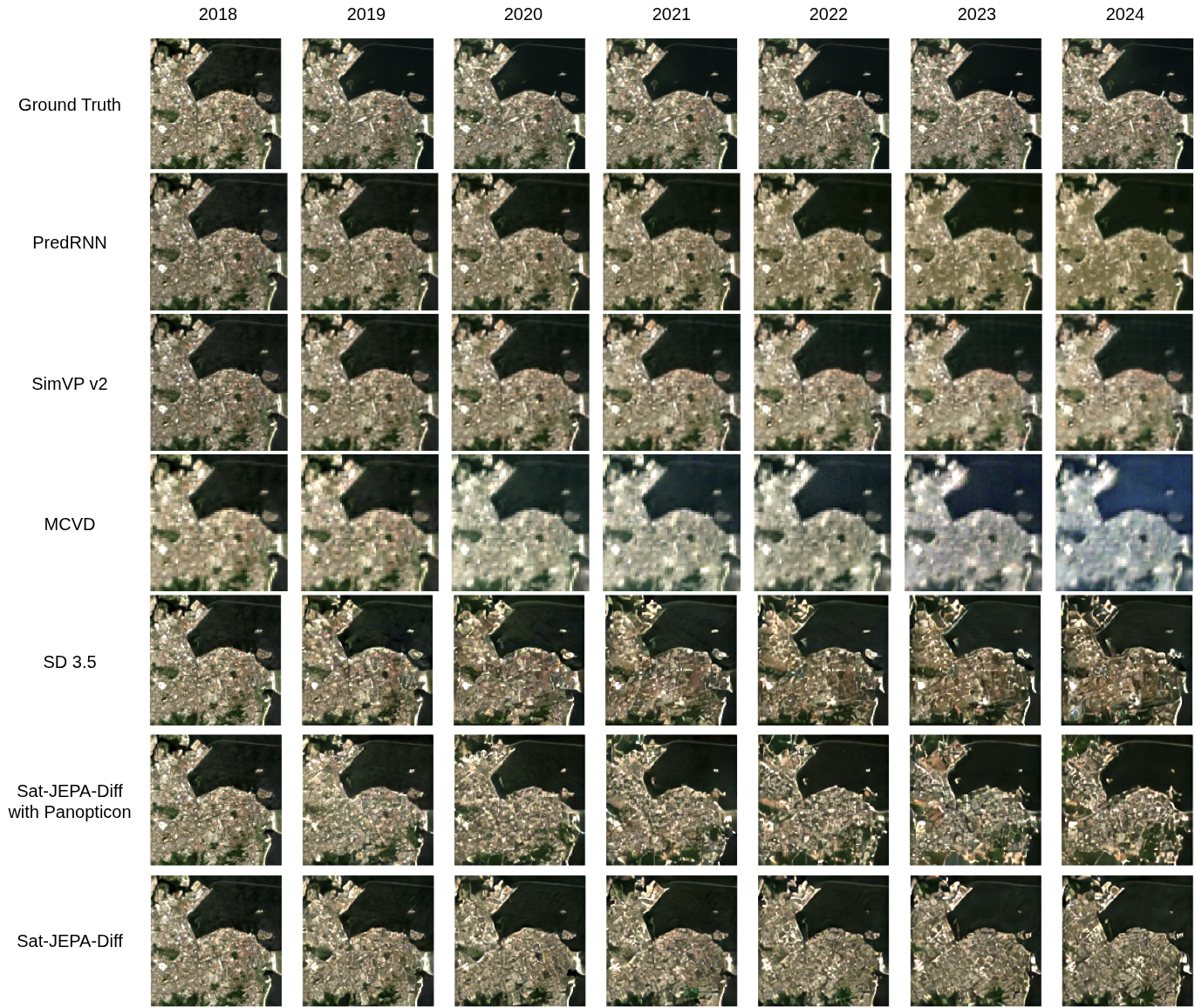}
\caption{Long-horizon autoregressive rollout comparison ($2018 \to 2024$) on the Rio de Janeiro Coast. \textbf{Top Row:} Ground Truth. \textbf{Rows 2-3:} Deterministic baselines rapidly degrade into spectral blurring (spatial collapse) after 2-3 steps. \textbf{Bottom Row:} Sat-JEPA-Diff maintains high contrast and structural sharpness throughout the 7-year horizon.}
\label{fig:rollout}
\end{figure}

To address the necessity of a multi-component loss formulation under our computational constraints, we performed an early trajectory ablation study (20 epochs) focusing on the IJEPA encoder and predictor operating on SOTA base model embedded vectors.

As shown in Figure \ref{fig:loss_ablation}, while the constrained 20-epoch window naturally allows the curves to converge to a similar limited region, early learning dynamics reveal critical differences. Removing certain adjustment terms (e.g., spatial variance or contrast loss) significantly delays the initial alignment with the target embedded vectors and destabilizes the early learning phase. This early trajectory analysis supports our reliance on full hybrid formulation to provide rapid and robust semantic orientation. A comprehensive full-scale ablation of 100 epochs with a diffusion backbone remains a key direction for future extended studies.

\subsection{Reproducibility}

\textbf{Random Seeds.} To enable CUDA's determinism, we lock the NumPy and PyTorch seed values to 0 and set \texttt{torch.backends.cudnn.benchmark = True}.

\textbf{Data Split.} The dataset is randomly shuffled before being divided into 80\% training and 20\% validation subsets.

\textbf{Code Availability.} We built this pipeline on PyTorch 2.0+, using PEFT for Diffusers for SD3.5 and LoRA integration. The YAML configuration files track all experimental settings.

\section{Autoregressive Stability Analysis}\label{app:rollout}

We assessed long-term stability using an autoregressive rolling method. Starting with the actual image from 2018, we generated predictions up to 2024 by feeding previous outputs back into the network. Figure \ref{fig:rollout} shows how deterministic methods (PredRNN, SimVP) rapidly blur high-frequency details to minimize MSE. Sat-JEPA-Diff successfully avoids this spatial collapse, preserving clear textures and sharp boundaries across all time steps.

\end{document}